\documentclass[lettersize,journal]{IEEEtran}
\usepackage{cite}
\usepackage{amsmath,amssymb,amsfonts}
\usepackage{algorithmic}
\usepackage{graphicx}
\usepackage{textcomp}
\usepackage{cite}
\usepackage{url}
\usepackage{hyperref}
\usepackage{stfloats}
\usepackage{verbatim}
\usepackage{xspace}
\usepackage[linesnumbered,ruled,noend]{algorithm2e}
\usepackage{color}
\usepackage{tabu}
\usepackage{tabulary}
\usepackage{pifont}

\usepackage{xcolor, colortbl}
\usepackage[colorinlistoftodos]{todonotes}
\SetAlFnt{\small}
\SetAlCapFnt{\large}
\SetAlCapNameFnt{\large}
\usepackage{algorithmic}

\usepackage{multirow}
\usepackage{array}

\usepackage{subcaption}
\usepackage{tabularx}
\usepackage{adjustbox}
\usepackage{booktabs}
\usepackage{array}

\def\name{{EMERSK}\space}
\def\aer{{automatic emotion recognition}\space}

\begin{document}

\title{\name-Explainable Multimodal Emotion Recognition with Situational Knowledge}

\author{Mijanur Palash and~Bharat Bhargava,~\IEEEmembership{Fellow,~IEEE}
\IEEEcompsocitemizethanks{\IEEEcompsocthanksitem The authors are with the Department
of Computer science, Purdue University,
USA, 47906.\protect\\

E-mail: (mpalash@purdue.edu,bbshail@purdue.edu)
 }}



\maketitle


\begin{abstract} 

Automatic emotion recognition has recently gained significant attention due to the growing popularity of deep learning algorithms. One of the primary challenges in emotion recognition is effectively utilizing the various cues (modalities) available in the data. Another challenge is providing a proper explanation of the outcome of the learning. To address these challenges, we present Explainable Multimodal Emotion Recognition with Situational Knowledge (EMERSK), a generalized and modular system for human emotion recognition and explanation using visual information. Our system can handle multiple modalities, including facial expressions, posture, and gait, in a flexible and modular manner. The network consists of different modules that can be added or removed depending on the available data. We utilize a two-stream network architecture with convolutional neural networks (CNNs) and encoder-decoder style attention mechanisms to extract deep features from face images. Similarly, CNNs and recurrent neural networks (RNNs) with Long Short-term Memory (LSTM) are employed to extract features from posture and gait data. We also incorporate deep features from the background as contextual information for the learning process. The deep features from each module are fused using an early fusion network. Furthermore, we leverage situational knowledge derived from the location type and adjective-noun pair (ANP) extracted from the scene, as well as the spatio-temporal average distribution of emotions, to generate explanations. Ablation studies demonstrate that each sub-network can independently perform emotion recognition, and combining them in a multimodal approach significantly improves overall recognition performance. Extensive experiments conducted on various benchmark datasets, including GroupWalk, validate the superior performance of our approach compared to other state-of-the-art methods.

\end{abstract}



\section{Introduction}

\IEEEPARstart{E}{motion} shapes our social life by influencing our communications with others. Therefore, automatic human emotion recognition (ER) holds significant potential in various aspects of our lives. In the current era of online learning, which has become prevalent due to the Covid-19 pandemic, an integrated ER system can help teachers maintain an effective learning environment by providing insights into the emotional state of students. Similarly, a car equipped with driver emotion recognition capability can proactively prevent road rage or accidents by alerting the driver when they are tired, frustrated, or angry. Furthermore, the implementation of an ER system in CCTV cameras can enable the detection of individuals displaying anger near sensitive locations such as schools or children's playgrounds, triggering timely alarms to prevent potential harm, including deadly school shootings. Human-computer interactions, law enforcement and surveillance, interactive games, consumer behavior analysis, customer service enhancement, and healthcare are just a few examples of the diverse fields where emotion recognition technology can significantly impact outcomes and experiences.

 
Moreover, we perceive other people's emotions using both visual and non-visual cues. Visual cues include facial expressions, posture, gestures, eye movement, and walking gait, to name a few. Non-visual cues encompass speech, text, brain signals, and EEG signals, among others. Working with visual cues is more common than working with non-visual cues, as visual cues are more easily obtainable. Facial expressions and postures can be observed directly by looking at a person or analyzing images or videos captured by regular cellphones, CCTV cameras, or similar devices. However, the same convenience does not apply to non-visual cues. For instance, obtaining a brain scan requires specialized instruments to be attached to the individual, and obtaining permission for such procedures can be challenging. Moreover, the knowledge of them being recorded can potentially influence the subject's emotional state. Therefore, this work primarily focuses on visual cues, specifically facial expressions, postures, and gaits, for the purpose of emotion recognition.

\begin{figure}[!t]
 \centering
\subfloat[]
{\includegraphics[width=0.4\linewidth]{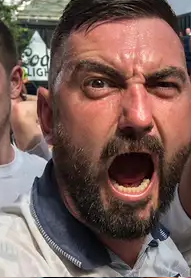}}
\hfill
\subfloat[]
{\includegraphics[width=0.545\linewidth]{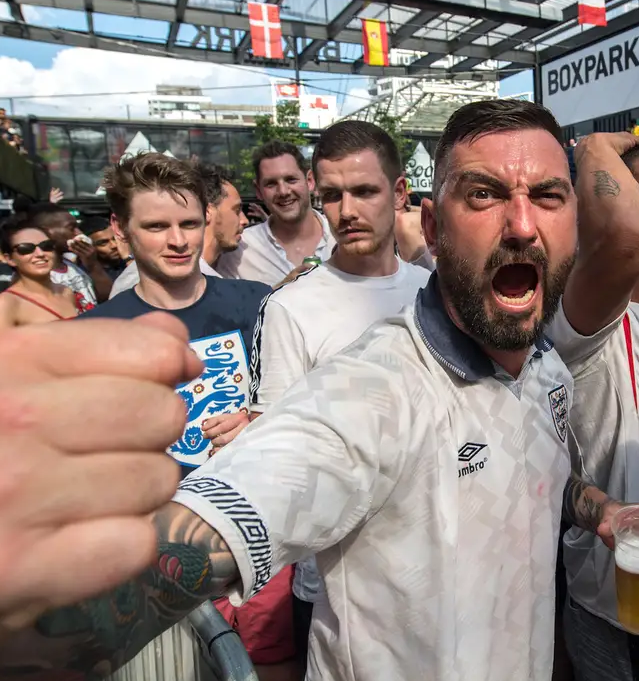}}
\caption{Importance of context in emotion recognition. (a) Without background the subject appears angry. (b) The background context shows the subject is happy and celebrating the win of his favourite soccer team with friends.}
\label{sf}
\end{figure}

Many of the existing works use only one type of cue (unimodal) such as facial expression~\cite{FER2013,mollahosseini} or gait~\cite{randhavane2019identifying} etc. However, solely depending on a single mode can make the model less reliable in wild deployment. For example, a facial expression model trained on many of the existing benchmark datasets with no masked sample will perform poorly when encountering a subject wearing mask, which is very common during the Covid-19 pandemic. Similarly, a person\textquotesingle s body may be blocked from the camera view by an obstacle and we may not have posture or gait information. So considering multiple modes at the same time can make the model more reliable. Moreover, several prior works show that combining multiple cues (multimodal) can results in higher accuracy in \aer \cite{d2010multimodal, emoticon}. A person with a smiley face (mode 1) is possibly happy, but if we know they have open arms and stand straight (mode 2) we can be much more confident in our deduction. 

 Likewise, it is well accepted that situation around a person plays a non-trivial role in shaping their general feelings~\cite{situational}. Deciding emotional state may prove inadequate or wrong if we do not consider the situation around the subject. For example, in the cropped photo at figure~\ref{sf}(a), the subject appears as an angry person. But the background in the uncropped photo in figure~\ref{sf}(b) gives us more information about the situation. We see the England national team fans celebrating wildly when the team qualifies for the FIFA World Cup 2018 semifinals after 28 years, and they are extremely happy and excited. We see people around the subject laughing and relaxed which would be different around an angry person. Thus the background and the emotional status of other people around the subject also give us valuable clues. Moreover, people in a particular area at a particular time tend to have similar feelings as a group. We do not normally see people very happy at a funeral or entryway to a hospital.  When there is a goal in a soccer match, all the supporters of that team would be happy together while the opposition would be unhappy.  One recent research shows that the students experienced a two-day lift in their mood when the university\textquotesingle s football team was victorious\cite{nfl}.  Consequently, some of recent works showed improved performance by considering background contextual information \cite{emoticon,kosti2019context,lee2019context}. 
 
Explainable artificial intelligence~\cite{xai} deals with the transparency of AI-based systems. The lack of interpretability in black-box-style deep learning models has been known to create trust and reliability issues, necessitating a greater emphasis on explainability. Various methods have been proposed to generate explanations of deep learning models, including Saliency Map~\cite{saliency}, Grad-CAM~\cite{gradcam}, SHAP~\cite{shapley}, and LIME~\cite{lime}. To this end, in~\cite{xai1}, the authors proposed an explainable pipeline utilizing facial action units and an agnostic LIME model to visualize the active regions of the face. Additionally, in~\cite{xai2}, the authors employed Saliency Map and Grad-CAM methods to generate an explainable model for driver's facial emotion recognition. This growing body of research highlights the importance of explainable AI in ensuring the reliability and trustworthiness of AI-based systems.

In this work, we introduce EMERSK, an novel and modular system of emotion recognition that takes into account situational knowledge to generate explainable results. By leveraging multi-modal data and context from the background, EMERSK is able to classify emotions with greater accuracy. Additionally, the system creates situational knowledge by utilizing place categories, Adjective-Noun Pairs (ANP), and average emotion scores, which serves to enhance the interpretability of the results. Overall, the main contributions of this work are:

\begin{itemize}
\item We propose a modular architecture for emotion recognition from multiple cues, where each mode works independently on one cue and can be easily added or removed.
\item We offer a working prototype which uses face, posture and gait as cues and shows improved performance in comparison with prior works.
\item We investigate the trade-off between energy cost and performance gain in multimodal emotion recognition.
\item We introduce a new approach for constructing situational knowledge using individual modes classification, place type, Adjective-Noun Pairs (ANPs) extracted from the scene, and average emotion score.
\item We present a novel method for explainable emotion recognition by leveraging the generated situational knowledge.
\end{itemize}

The remainder of the article is structured as follows:  In section~\ref{rw} we discuss related works. Next, we describe our proposed architecture in section~\ref{pm}. Then we report our experimental results in section~\ref{er}. Finally, concluding remarks and future works are discussed in section~\ref{cfw}.

\section{Related Work}\label{rw}

Unimodal works deal with single data modalities. Among different modalities used in the literature, the facial expression is the most widely used one. Gan et al.~\cite{gan} achieved improved accuracy on FER-2013 data set using ensemble CNN and a label perturbation strategy. In~\cite{dhankhar}, authors used the ensemble method and transfer learning with VGG16 and RESENT-50 to overcome the limitations of the CNN-only methods. In~\cite{step, renda, amp-net,2021fer,ac, vua, li2021human,marrero2019feratt } authors proposed various methods to identify the emotion based on individuals walking face, posture and gait using different deep learning techniques including CNN, encoder-decoder network, self-attention, recurrent neural network (RNN) and long short-term memory (LSTM) etc. 

Apart from the uni-modal works, several researchers combined multiple modalities for automatic emotion recognition. In~\cite{sikka2013multiple}, authors combined multiple visual and audio features for \aer using kernel and Support Vector Machines (SVM) methods. In~\cite{gunes2007bi}, authors used the face and upper body gestures for multi-modal emotion recognition.  In \cite{scherer2007multimodal}, authors combined facial, vocal features, and body movements (posture and gesture) to discriminate 14 emotion categories. In~\cite{castellano2008automated}, authors used facial features, speech contours, and gestures. Similarly, in~\cite{chen2023coupled,poria2016,sun2018affect,mibdl,mittal2021,bhatia2021lstm } authors proposed various methods to identify the emotion based on multimodal data.

Recently some researchers proposed emotion recognition systems with context. In~\cite{emoticon}, authors leveraged psychological interpretation of context and used depth-based CNN to model the socio-dynamic interactions between the agents. In~\cite{kosti2019context}, authors used two-stream architecture where one stream takes only the visible part of the body of the subject, and the other takes the whole image as input. In~\cite{lee2019context}, they followed similar architecture with one stream focusing on the face of the subject while the other takes the whole image without the face to generate the context. In~\cite{meer}, authors combined cues from face, text and speech with multiplicative fusion. 

 Group emotion recognition deals with the emotion analysis of a group. Researchers in ~\cite{group1,group3,group} showed the interaction between individual and group emotions and proposed several methods to determine the group emotion from individual emotion using face, body and other cues. 
 
 The Places Database~\cite{places} is a collection of 10 million labeled scene photographs from the environment types encountered in the world. It also provides pre-trained scene classification CNNs (Places-CNNs) which can be used for scene category and attribute identification. In~\cite{anp}, the authors constructed a Visual Sentiment Ontology (VSO) database consisting of more than 3,000 Adjective Noun Pairs (ANPs). They also propose SentiBank, a visual concept detector library that can be used to detect the presence of various ANPs in any image.
 
 \begin{figure*}[b]
\centering
   \includegraphics[width=0.89\linewidth]{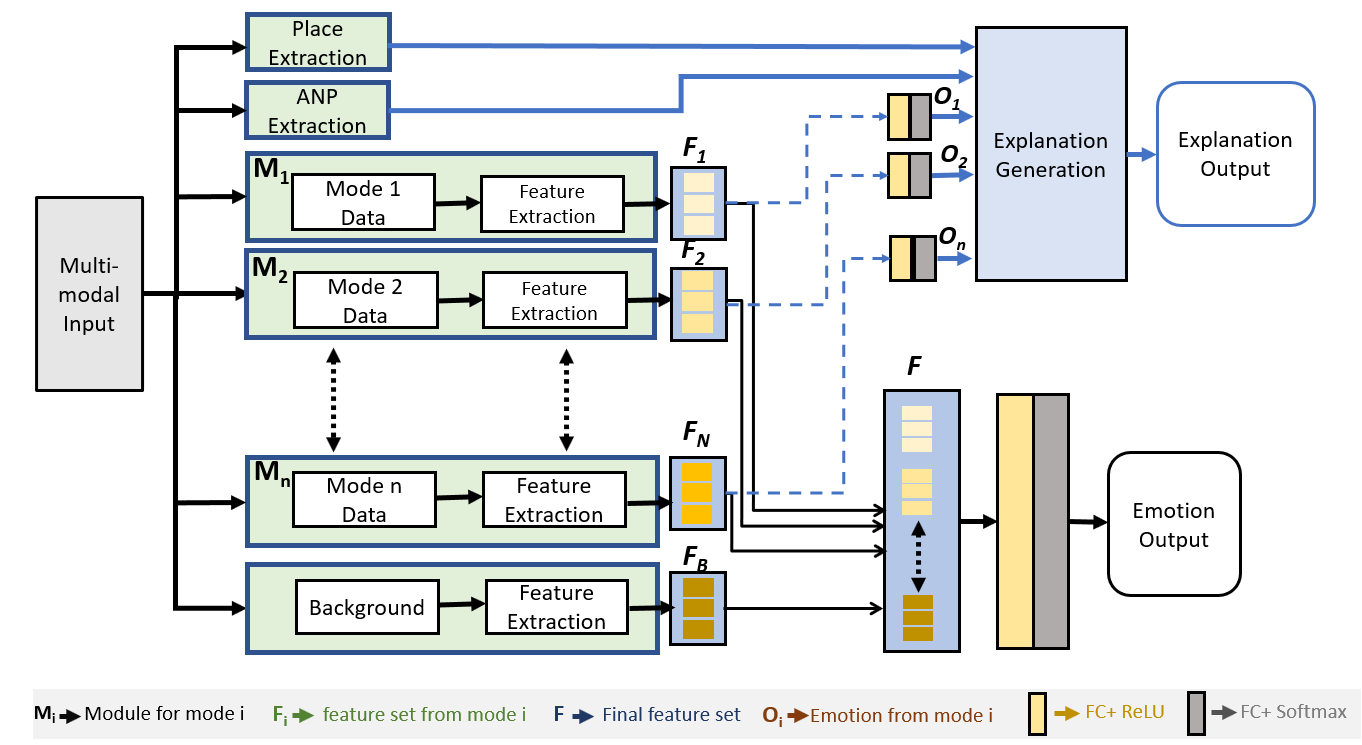}
   \caption{Proposed modular system architecture. $M$ represents the data processing and feature generation module, $F$ indicates the extracted feature set, and $O$ denotes the individual modular classification output. The subscript $i$ indicates the specific data mode, such as face or gait. }
   \label{fig:sysD}
\vspace{-0.1in}
\end{figure*}

\section{Proposed Method: \name}~\label{pm} In this section we discuss different parts of our  proposed modular architecture. From the input data $x \in (I,V)$, where $I$ represents image type and $V$ represents video type data we collects available data modes $m\in (m_1, m_2,..., m_n)$. The possible modes can be the facial expression, posture, gait, background etc. as discussed earlier. However other modes such as voice or text can be used similarly. Each mode $m_i$ is processed by corresponding module $M_i$. During training time each module is trained to classify emotion from that particular data mode only. However, in the multimodal situation, we simultaneously classify emotions and also collect deep feature representation $f_i$ for that mode.


All the features from each module are fused to generate the final feature set $F\in(f_1,f_2,...,f_n)$. This final feature set $F$ is passed through a classification network to classify them into $k$ groups of emotions. Similarly, individual module feature sets ($f_i$) are also passed through separate classification networks to generate emotion classification ($O_i$) based on that module only. The resulting final emotion from the multimodal operation can be compared with the individual modular results $(O_i)$ for explanation generation (section~\ref{exp}) or anomaly detection (section~\ref{anomaly}).

This variable modular architecture provides flexibility. For example, to detect the emotion of students in a Zoom class, we can get the face, upper-body posture, voice etc. but not gait. Similarly, a system which monitors the emotions of the crowd using CCTV footage would not get voice or brain scan information. Moreover, in both of these situations, where we are working with live videos, we would have the subject\textquotesingle s background and situational knowledge. Our target in this work is to provide a general architecture of emotion recognition which can work with most types of data for different application scenarios. With a flexible modular architecture proposed here, it is possible further fine-tune the model based on the available data modules and specific task requirements.


For the remaining part of this work, we would only use face, posture, gait, background and situational knowledge as our data modes to demonstrate the benefit of this modular architecture. These particular modes are selected due to the ease of training as they normally co-exist and are available in many datasets. Also, most emotion recognition tasks such as classroom or crowd situations mentioned above most likely would have data in these modes.

\subsection{Module 1: Face}
The face module deals with the face area of the input data. It performs face detection, face deep feature generation and classification of emotion from the face data.

\subsubsection{Face Detection and Cropping}\label{fd}

For face detection, we leveraged the technique outlined in the works of Bazarevsky et el.~\cite{blazepose}. It detects the face area using 6 landmarks on the face. It also supports the detection of multiple faces from an image.  Blazeface uses a very compact feature extractor convolutional neural network designed specifically for lightweight but accurate face detection.

\begin{figure*}[]
\centering
   \includegraphics[width=.99\textwidth]{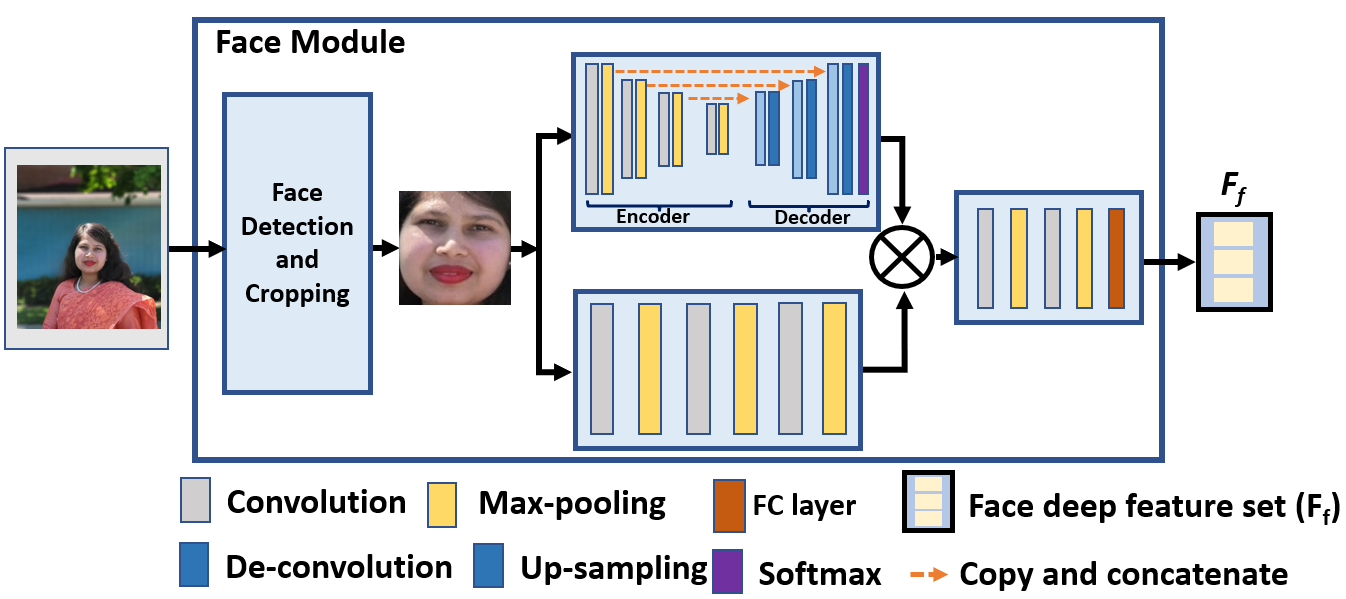}
   \caption{\name face module.}
   \label{cnn}
\vspace{-0.1in}
\end{figure*}


\subsubsection{Facial Feature Extraction}

We employed a two-stream network for face feature extraction, as depicted in Figure \ref{cnn}. In the lower stream, a convolutional neural network (CNN) is used as a feature extractor. Each input image is treated as a tensor of size $C \times W \times H$, where $C$, $W$, and $H$ represent the number of channels, width, and height, respectively. The 2D convolution and polling kernels can be represented as a tensor of size $k \times k$. We employed a network with 3 convolutions and 3 max-pooling layers using a convolution kernel of dimensions $3 \times 3$ and a pooling kernel of $2 \times 2$ with stride 1 and necessary padding.

The upper stream is an encoder-decoder style attention module. We constructed our encoder-decoder network by following the methods used in \cite{unet,segnet,deconv,santhoshkumar2020vision} in image segmentation. The encoder part comprises of four $3\times 3$ convolution layers with ReLU and $2\times2$ max-pooling layers. The decoder part utilizes layers of $2\times2$ up-sampling, de-convolution~\cite{deconv}, and ReLU. The corresponding layers of the encoder and decoder are joined with copy and concatenation links. The up-sampling layer utilizes memorized max-pooling indices from the corresponding encoder and generates a sparse feature map. The deconvolution layers enlarges and densifies the sparse feature map through convolution-like operations with trainable decoder filter banks. This encoder-decoder based attention mechanism has demonstrated improved performance in recent works in facial expression recognition such as \cite{marrero2019feratt}. The encoding segment converts the input data into a lower-dimensional feature space and the decoding segment reconstructs the original input from the encoded features. The attention mechanism allows the decoder to selectively attend to certain parts of the input image. During our experiments, we observed that the two stream architecture outperforms the architectures based solely on CNN.

The resulting output from the two streams was multiplied and passed through another two convolution and maxpooling layers. One fully connected (FC) layer was added after the convolutional layers. For multimodal operation, we collected the face deep feature set $(F_P)$ from this FC layer. However, for stand-alone operation, two additional FC layers were used for training, and cross-entropy loss was utilized as the training loss function.

\subsection{Module 2: Posture}
\begin{figure}[]
\centering
   \includegraphics[width=1.0\linewidth]{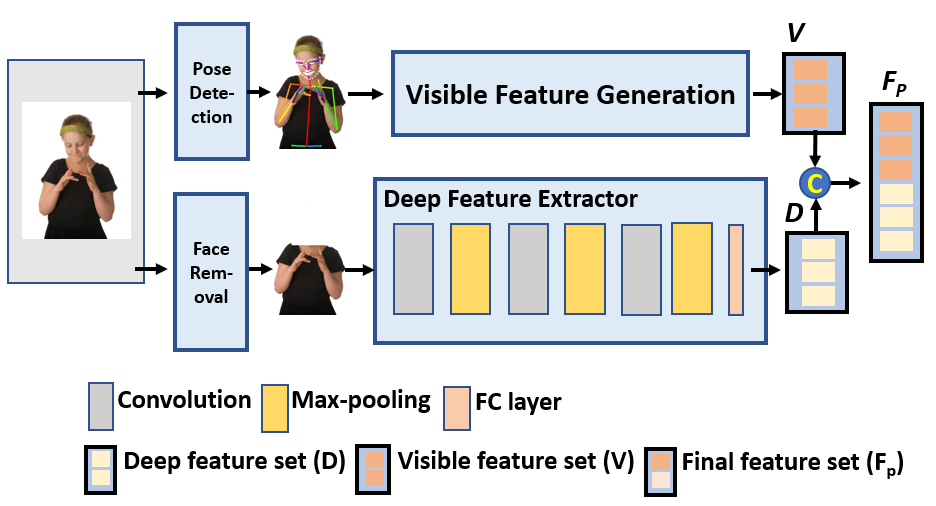}
   \caption{\name posture module.}
   \label{fig:openpose}
\vspace{-0.1in}
\end{figure}

The term `posture' refers to our physical form, the way we hold ourselves during activities like sitting, standing or sleeping etc. Posture provides a plethora of information about our feelings and emotions. The posture module is responsible for emotion recognition from the posture data. The architecture of posture module is shown in figure~\ref{fig:openpose}.

\subsubsection{Posture Detection}



\begin{figure}[]
\centering
   \includegraphics[width=.8\linewidth]{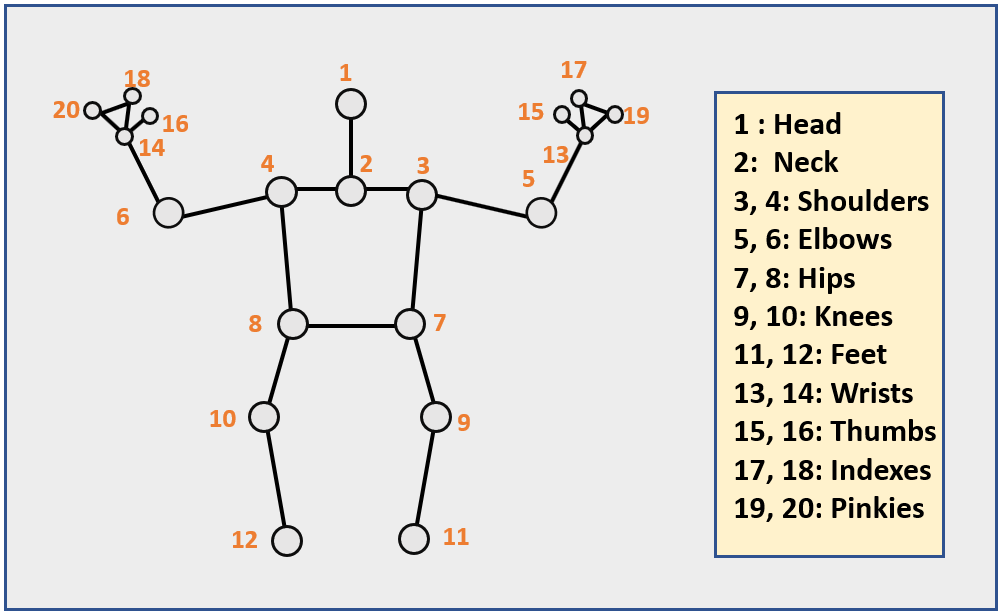}
   \caption{Kinematic representation of a human body.}
   \label{fig:body}
\vspace{-0.1in}
\end{figure}

We need a proper model of the human body to identify posture correctly. In this work, we use the kinematic approach to human body representation. Our model is 2D with 20 different body points as shown in figure~\ref{fig:body}. We use BlazePose~\cite{blazepose} to find these 20 body points. BlazePose provides human pose tracking using machine learning (ML) and offers 2D landmarks of a body from a single RGB frame. For each video frame, BlazePose provides the location of the body points in the frame. This information then can be used to generate more data regarding the pose. 


\subsubsection{Pose Visual Feature Set}
Following the works of the authors in~\cite{randhavane2019identifying} and~\cite{crenn}, we extract 24 features from the locations of the body points. These features $f_0, f_1, \dots , f_{23}$ are listed in the table~\ref{post_features}. They are in the form of volume, angle, distance and area. In~\cite{crenn}, they only used upper body features such as the area of the triangle between hands and neck, distances between hand, shoulder, and hip joints, and angles at neck and back. However, they overlooked the lower body such as foot joints, which also can convey important information, particularly while walking. In ~\cite{randhavane2019identifying}, they added areas, distances, and angles of the feet joints in the posture feature formulation. Finally a posture visual feature vector $V\in f^{24}$ is created using these features $f$ for each subject in a frame.

\subsubsection{Deep Feature Generation}
We also generate a deep feature set $D \in f_d^{100}$ from the RGB input image beside the calculated visible feature set $V \in f^{24}$. From our experimental results, we found that adding deep features with visible features resulted in increased accuracy due to CNN's ability to capture subtle points not captured by the visible feature. To generate these deep features $f_{d1}, f_{d2}, \dots, f_{d99}$ we first extract the rest of the body from the RGB image and also remove the face from it. For deep feature extraction, we use a CNN as shown in figure~\ref{fig:openpose}. For each input image, we collect the weights  $w_{1}, w_{2}, \dots, w_{99}$  at the output of the first fully connected layer.

\subsubsection{Final Posture Feature Set}
Final posture feature set $F_P$ is created by concatenating the visible $(V)$ and the deep ($D$) feature sets ($F_P=[V,D]$). All the feature values are normalized to be bounded between 0 and 1 ($0 \leq f \leq 1$). This feature set $F_P$ is used for multi-modal emotion recognition.
However, for the stand-alone emotion classification task from posture only, we train a deep neural network (DNN) with 2 fully connected layers and the cross-entropy loss function. This network takes the posture feature set $F_P$ and outputs the probability of each emotion class after a soft-max operation.

\begin{table}[]
\centering
\begin{tabular}{|c|c|}
\hline
\textbf{Feature Type}  & \textbf{Feature Description}  \\ 
 \hline
 
   The angle at& Neck by both shoulders\\
 
   &Right shoulder by
neck and left shoulder\\

&Left shoulder by
neck and right shoulder\\

&Neck by vertical and back\\

&Neck by head and back\\

&Right shoulder by the right arm and neck\\

&Left shoulder by the left arm and neck\\

&Right elbow by the right arm and right forearm\\

&Left elbow by the left arm and left forearm\\

&Hip between y-axis and torso\\
\hline
Distance between&
Right hand
and the hips\\

&Left hand
and the hips\\

&Right foot
and the hips\\

&Left foot
and the hips\\

&Right hand and the right shoulder\\

&Left hand and the left shoulder\\

&Right elbow and the hips\\

&Left elbow and the hips\\
\hline

Area of the triangle by&
Both
hands and neck\\

&Both shoulders and neck\\

&Both feet and the hips\\

&Both hands and hips\\

&Both elbows and neck\\
\hline
  Space taken & Bounding box volume\\
  \hline

\hline
\end{tabular}
\caption{Posture visual features.}
\label{post_features}
\end{table}

\subsection{Module 3: Gait}
Gait is defined as the way a person walks. Walking is a complex movement which requires balance and coordination of multiple body parts such as the brain, bones, muscles, heart, lungs etc. Our gait conveys meaningful information regarding our emotions. Our two streams gait module architecture is shown in figure~\ref{3dcnn}.

\subsubsection{Gait Deep Features}

\begin{figure}[]
\centering
   \includegraphics[width=.89\linewidth]{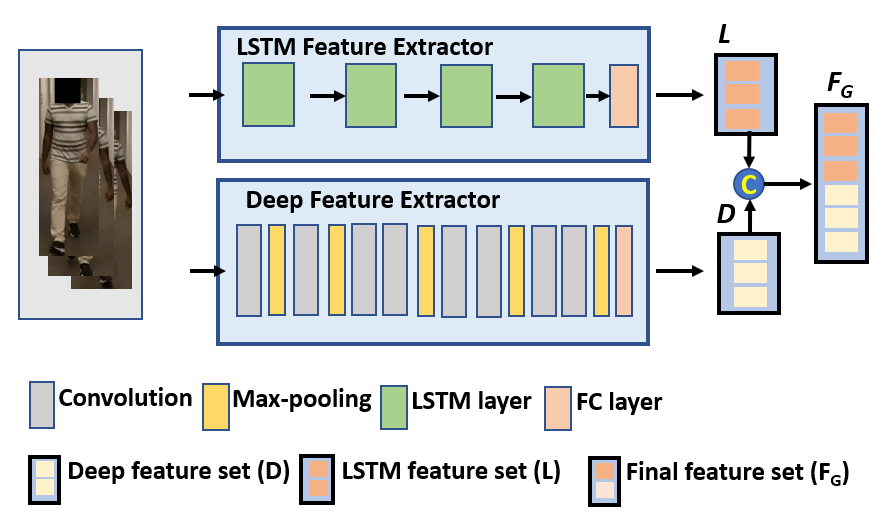}
   \caption{\name gait module.}
   \label{3dcnn}
\vspace{-0.1in}
\end{figure}

We take inspiration from ~\cite{3dcnn} and build a 3D CNN architecture for learning gait deep features from a video segment. A 3D CNN can capture temporal features as well. Each input video segment has 16 frames. We can represent the video segment as a tensor of size $C \times W \times H \times N$, where $C$ is the number of channels, $W$ and $H$ are width and height respectively and $N$ is the number of frames in the segment. The 3D convolution and polling kernel can be similarly represented as a tensor of size $d\times k \times k$, where $d$ is kernel depth and $k$ is the spatial size. We used a network with 8 convolutions, 5 max-pooling and 2 fully connected layers as shown in the figure~\ref{3dcnn}. We use convolution kernel of dimension $3 \times 3\times 3$ and pooling kernel of $2 \times 2\times 2$ (except the first one which has pooling kernel size $1 \times 2 \times 2$) with stride 1 and necessary padding in both spatial and temporal domain. RGB frames with the subject only while face and background removed are provided to the network. 

We also use Long Short Term Memory (LSTM) network following the work of~\cite{lstm}. This network can model long-term temporal dynamics as well as learn deep representations. We build an LSTM network by stacking 4 LSTM layers as shown in the figure. Same as the 3D CNN part, the RGB frame sequence with the body (background and face removed) is fed into the LSTM network and deep features are collected from the last layer.

\subsubsection{Gait Final Feature Set}
The final deep feature vector is a concatenation of 3D-CNN and LSTM features $F_G=[D, L]$
For the stand-alone classification part, we used a DNN with 2 fully connected layers and the cross-entropy loss function on $G$. For multi-modal training, deep feature set $F_G$ is used directly.

\subsection{Background Context}
\begin{figure}[]
\centering
   \includegraphics[width=.99\linewidth]{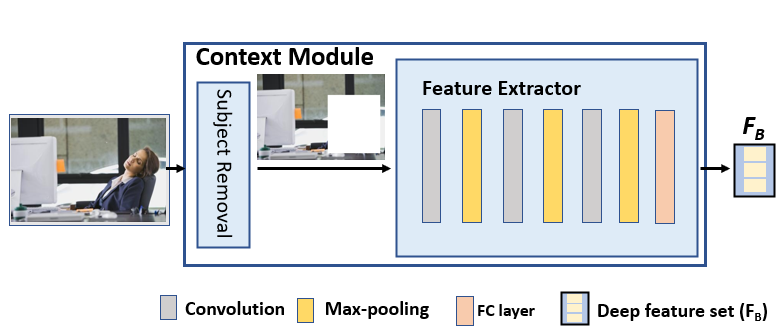}
   \caption{\name background context module.}
   \label{context}
\vspace{-0.1in}
\end{figure}

The architecture of background contextual feature extraction module is shown in figure~\ref{context}.
Contextual information from the scene background helps to make more accurate recognition. To generate background contextual information we first remove the subject face and body from the image. The image is then passed through a CNN based feature extractor network to generate the deep feature set $F_B$.

\subsection{Final Feature Set}
Final features set ($F$) is the collection of features from each modules. We concatenate the output feature sets from face ($F_F$), posture ($F_P$), gait ($F_g$
) and background ($F_B$) to get the final feature set $F$.
\[F=[F_F,F_P, F_G, F_B]\]

This final feature set $F$ is passed through two FC layers for multimodal classification. During training, we use cross-entropy as the loss function.

\subsection{Explainable Situational Knowledge Generation}
\subsubsection{Place and ANP}
It has been shown in the literature that the distribution of emotion varies significantly among different place categories. For example, people in a `ski\_slope' frequently appear happy while people in `working\_environments' appear sadder. Similarly, ANPs such as `young\_couple' is associated with happiness. This place and ANP information are used to generate an explanation for the result.  We use pre-trained AlexNet provided with the Places dataset~\cite{places} which gives place categories such as `classroom' and attributes such as `no\_horizon' and `enclosed\_area' etc. We also use Sentibank Adjective-Noun Pair (ANP) detectors~\cite{anp} which give ANPs like `young\_couple' and `outdoor\_wedding' from images.

\subsubsection{Average Emotion}\label{avgEmo}
Average emotion is collected in a group setting where we have multiple subjects in the same frame or frames from the same geographical area (spatial), bounded by a threshold distance $t_l$ and, within a short time interval (temporal), bounded by a threshold time difference $t_d$. For example, timestamped videos from multiple CCTV cameras around a complex can be used for each other. For each time interval $T$ we collect numbers of emotions recognised from each emotion category $N_e$ where $e \in (happy, sad, neutral,  angry)$. We maintain a 2-D list of average emotion scores where each row is time and each column represents one emotion category $e$. Value of each list entry is calculated by \[l_e=  \frac{N_e}{\sum_{e}(N_e)}\]
Together with place attributes and ANPs, average emotion can give us more explanation for the emotion.

\section{Experimental Results}\label{er}

In this section, we will be discussing our experimental results. We will first analyze the performance of each modality separately, and then we will present the overall multimodal performance. To avoid slowing down the multimodal operation, we refrained from using complex architecture for individual modules. Consequently, the stand-alone operation of individual modules may not always outperform the best results reported in the literature generated from complex models. Our aim is to achieve improved performance with simpler modules operating independently while achieving a greater improvement in the multimodal setting.

All experiments were conducted on a server PC with a 20-core 2.6 GHz CPU, 96 GB of available memory, and three GPU cards, each with 24.5 GB memory. To speed up computation, we used Pytorch multiprocessing and automatic mixed precision libraries. In cases, where  no working implementation of a particular method is available, we used our best effort implementation of that method.  We also selected datasets which are frequently used by recent works to allow for a direct comparison of our results with them.

\begin{table*}[]
\centering

\begin{tabular}{|c|c|c|c|c|c|c|}
\hline
 \textbf{Name}&\textbf{No of Items}   &\textbf{Type}&\textbf{ Data Modes} &\textbf{Setting} &\textbf{Situational Intelligence}  &\textbf{Classes}   \\ 
 \hline
   FER-2013\cite{FER2013}& 32,298 &   Image& Face & Posed& No& N, S, H, A, Sr, F and D  \\
  Emotic\cite{kosti2019context} & 23,571  &   Image& Face and Posture& Wild&Background  &N, S, H, A, Sr, F, D and 19 other classes   \\
   CAER-S\cite{caers} & 70,000     & Image& Face and Posture& TV shows&Background&N, S, H, A, Sr, F and D\\
   CAER\cite{caers}& 13201 & Video&Face and Posture& TV shows&  Background &N, S, H, A, Sr, F and D\\
  FABO\cite{gunes2007bi} & 206     & Video& Face and Posture&Posed& No&N, S, H, A, Sr, F, D, B, P and Ax\\
  EWalk\cite{randhavane2019identifying}&1384&Video& Posture and Gait&Posed& Background&N, S, H and A\\
GroupWalk\cite{emoticon}&45&Video&Face, Posture and Gait&Wild&Background and Group& N, S, H and A\\


\hline
\end{tabular}

\caption{Summary of the various emotion recognition datasets, with
N representing Neutral, S representing Sadness, Sr representing Surprise, H representing Happiness, F representing Fear, A representing Anger, B representing Boredom, P representing Puzzlement, Ax representing Anxiety, C representing Contempt, and D representing Disgust.}
\label{datasets}
\end{table*}


\subsection{Datasets}


In the present body of literature, a plethora of datasets have been proposed for the task of emotion recognition. Table \ref{datasets} presents a comprehensive summary of the datasets employed in our work, highlighting their unique characteristics. For instance, FER-2013~\cite{FER2013} solely consists of facial images, without any accompanying information on posture, gait, or situational context. The Emotic~\cite{kosti2019context} dataset, on the other hand, contains images with the face, upper body, and background. The CAER and CAER-S~\cite{caers} datasets encompass video clips and images from TV shows, primarily focusing on upper body and facial images with background. The FABO~\cite{gunes2007bi} dataset comprises acted emotion videos depicting the face and upper body against a green screen. The EWalk~\cite{randhavane2019identifying} dataset is composed of acted videos featuring walking individuals with gaits and annotated emotions. Lastly, the GroupWalk~\cite{emoticon} dataset contains 45 videos of real-world scenarios, encompassing face, posture, and gait information. This dataset also provides situational intelligence in the form of background and other people's emotions.

\subsection{Face Emotion Recognition}

We leveraged the images from the datasets enlisted in Table~\ref{datasets} to train our facial expression recognition module. Subsequently, we evaluated its efficacy in recognizing emotions from facial expressions by benchmarking its accuracy against various state-of-the-art approaches proposed in the recent literature. The test accuracy of our method is computed as follows:

\[Accuracy=\frac{\#N_c}{\#N_t}\]

Where, $\#N_c$ indicates the number of items correctly predicted and $\#N_t$ indicates the total number of items in the test dataset. 

Our experimental results demonstrate that our proposed method achieves a test accuracy of 78.62\% on the FER-2013 dataset. Our approach exhibits superior performance in comparison to several other contemporary works for facial emotion recognition, as evidenced by the results presented in Table~\ref{t_acc1}. The closest performance in terms of accuracy was achieved by AMP-Net~\cite{amp-net}. Our approach outperforms them by a margin of 
4.14\%.

\begin{table}[h]

\centering
\begin{tabular}{|c|c|c|}
\hline
\textbf{Author}&\textbf{ Year}   &\textbf{Accuracy(\%)}  \\ 
 \hline
   Renda et al.\cite{renda}& 2019 & 71\\
   Gan et al.\cite{gan}&2019&   73.73\\
   Joseph et al.\cite{2021fer}&2021&67\\
   A-C\cite{ac}&2022&72.03\\
   AMP-Net\cite{amp-net}&2022&74.48\\
  \textbf{ \name}&\textbf{2023}& \textbf{78.62}\\
\hline
\end{tabular}
\caption{Comparison between our proposed \name face module and other recent works on the FER-2013 dataset. }
\label{t_acc1}

\end{table}

\begin{figure}[]
\centering
   \includegraphics[width=.99\linewidth]{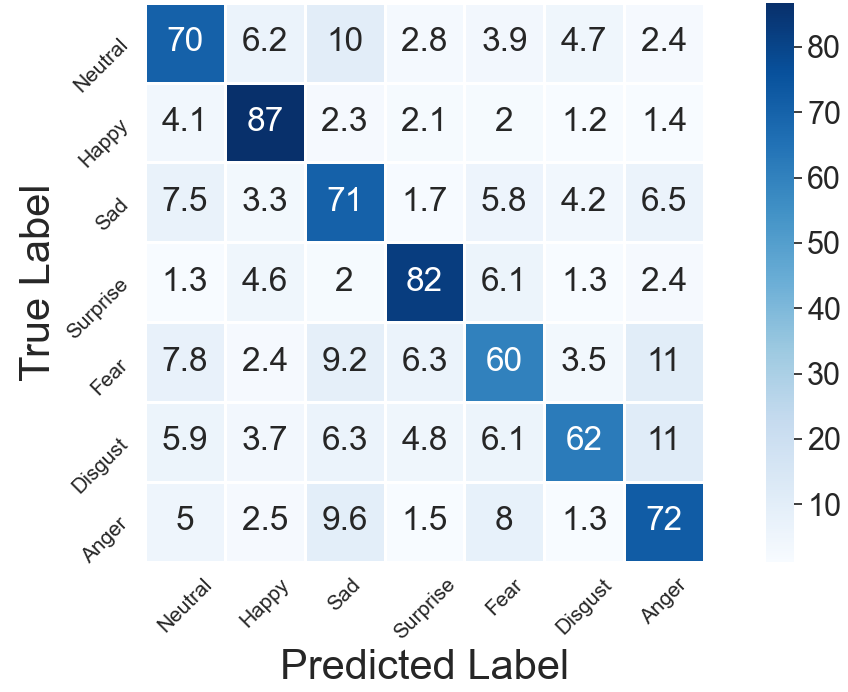}
   \caption{Confusion matrix on the FER-2013 dataset: the vertical axis shows the actual emotion labels, while the horizontal axis shows the predicted ones. Each entry of row i and column j indicates the percentage of samples with the true label of row i and the predicted label of column j. The intensity of the color indicates the magnitude of the percentage, with darker colors indicating higher values.}
   \label{conf_mat}
\vspace{-0.1in}
\end{figure}

Figure~\ref{conf_mat} shows the confusion matrices of our model for the FER-2013 dataset. From the confusion matrix of FER-2013, we can see some classes such as `Happiness' and `Surprise' are better recognizable while `Disgust' and `Fear' classes are not.

In addition, we also compare the result of our method with several recent works on the CAER-S dataset, as tabulated in Table~\ref{result_caers}. Notably, our approach outperforms the closest result reported by Li. et al.~\cite{li2021human} by a margin of 7.58\%. A summary of emotion recognition performance on various benchmark datasets using EMERSK face module is shown on Table~\ref{result_datasets}

\begin{table}[]
\centering
\begin{tabular}{|c|c|c|}
\hline
\centering
 \textbf{Method} &\textbf{Year}&\textbf{Test Accuracy(\%)}    \\ 
 \hline
  Lee et al.~\cite{lee2019context}&2019& 73.51 \\
   Kosti et al.~\cite{kosti2019context}&2019& 74.48 \\
   Li et al.~\cite{li2021human} &2021& 84.82 \\
   \textbf{\name}&\textbf{2023}&\textbf{92.4}\\

\hline
\end{tabular}
\caption{Comparison between our proposed \name face module and other recent works on the CAER-S dataset.}
\label{result_caers}
\end{table}

\begin{table}[]
\centering
\begin{tabular}{|c|c|}
\hline
\centering
 \textbf{Dataset} &\textbf{Test Accuracy(\%)}    \\ 
 \hline
  CK+& 98.6 \\
   FER-2013& 78.62 \\
   AffectNets&71.5 \\
   CAER-S & 92.4\\
   FABO & 98.1    \\

\hline
\end{tabular}
\caption{Accuracy of emotion recognition on various benchmark datasets using \name face module.}
\label{result_datasets}
\end{table}

\subsection{Posture Emotion Recognition}

For training and evaluating our posture module, we employed the FABO dataset, and achieved a test accuracy of 78.1\%. Additionally, we investigated bimodal emotion recognition by combining facial expression and posture information, following prior research. The results of this experiment are presented in Table~\ref{p_fabo}. Notably, while our posture module alone may not be as effective as some of the more complex approaches, our bimodal method surpasses them in performance. Specifically, our bimodal method outperforms the closest reported result by CMEFA~\cite{chen2023coupled} by a margin of 3.15\%.



\begin{table}[]
\centering

\begin{tabulary}{1.0\columnwidth}{|c|c|c|c|}
\hline
&&\multicolumn{2}{|c|}{\textbf{Accuracy (\%)}} \\\cline{3-4}
 \textbf{Author}&\textbf{ Year}   &\textbf{Posture}&\textbf{Bimodal}\\ 
   \hline
  
   CCCNN ~\cite{poria2016} &2016&74.8&93.65\\
   Sun et al.~\cite{sun2018affect} &2018&90.51&92.24\\
   MIBDL\cite{mibdl}&2021&79.71&92.54\\
CMEFA ~\cite{chen2023coupled}      &2023&\textbf{88.42}&93.58\\
   \name& 2023    &78.1&\textbf{96.73}\\ 
\hline
\end{tabulary}
\caption{Comparison between our proposed \name posture
module and other recent works on the FABO dataset.}
\label{p_fabo}
\end{table}

\begin{table}[]
\centering

\begin{tabulary}{1.0\columnwidth}{|c|c|}
\hline
 \textbf{Dataset}    &\textbf{Accuracy (\%)}  \\ 

\hline
FABO& 78.1\\
   Emotic   &81.2\\ 
   CAER-S&79.1\\
\hline
\end{tabulary}
\caption{Accuracy of emotion recognition on various
benchmark datasets using \name posture module.}
\label{p_other}
\end{table}

We further evaluated our posture module in two additional benchmark datasets, namely Emotic and CAER-S. The  results of the experiments on these datasets are shown in the table~\ref{p_other}. Our method achieved 80.15\% average test accuracy on these two datasets. 

\begin{table}[]
\centering

\begin{tabulary}{1.0\columnwidth}{|c|c|c|}
\hline
 \textbf{Author} &\textbf{Year}    &\textbf{Accuracy (\%)}  \\ 
 \hline

   Daoudi et al.~\cite{daoudi}&2017 & 42.52\\
  Li et al.~\cite{li18}&2018 &53.73\\
  Bhatia et al.~\cite{bhatia2021lstm}&2021&77.81\\
  \textbf{Tanmay et al.~\cite{randhavane2019identifying}} &\textbf{ 2022}&\textbf{ 80.07}\\
  \name &2023& 80.01\\
\hline
\end{tabulary}
\caption{Gait modules performance compared with other state of the art methods on EWalk dataset}
\label{g_other}
\end{table}

\subsection{Gait Emotion Recognition}
We conducted an evaluation of our gait-based emotion recognition modules using the EWalk dataset. Our model achieved a test accuracy of 80.01\% on this dataset. In Table~\ref{g_other}, we present the results from our study alongside those of other recent works.  Our approach exhibits improved performance in
comparison to several other contemporary works for gait
emotion recognition and closely matches the best performing works of Tanmay et al.~\cite{randhavane2019identifying}.

\subsection{Multimodal Emotion Recognition Results}
In this experiment, we tested overall performance of proposed \name system when all the modalities work together. For this, we use the GroupWalk dataset which has all the required modalities we need. Additionally, for better comparison with other state of the art works, we use the mean average precision (mAP) score as the performance metric in this dataset. mAP is a widely used evaluation metric which is a measures of the quality
of the algorithm. Average precision (AP) is the area under the precision-recall curve, which measures how well the algorithm retrieves relevant instances. The mAP is calculated as the average of the AP values for all the classes in the dataset. 

Our model achieved an mAP of 76.3 on this dataset. From the table~\ref{multi} it is clear that our model outperforms all other works in this dataset. This is due to our usage of all the three available modalities (face, posture and gait) with background context effectively. From the results, we can see our work outpeforms the closest performing work of Mittal et al.~\cite{emoticon} by an mAP value of 8.68.

\begin{table}[]
    \centering
    \begin{tabular}{|c|c|c|}
    \hline
     \textbf{Author} & \textbf{Year}&\textbf{Average Precision (mAP)}  \\
        \hline
        Kosti et al.~\cite{kosti2019context}&2019 & 58.42\\
        Lee et al. ~\cite{lee2019context}&2019& 62.58\\
        Mittal et al.~\cite{emoticon} &2020& 65.83\\
        Mittal et al. ~\cite{mittal2021}&2021&67.62\\
       \textbf{\name} &\textbf{2023}&\textbf{76.3}\\
        \hline
    \end{tabular}
    \caption{Comparison between our proposed \name multimodal emotion recognition and other recent works on the GroupWalk dataset.}
    \label{multi}
\end{table}



In additional study, we evaluated the efficacy of our multimodal approach using GEMEP~\cite{banziger2010introducing}, another benchmark dataset. The GEneva Multimodal Emotion Portrayals (GEMEP) has video recordings of actors portraying various emotional states. It has annotated face and full body expressions of 18 emotions from 10 subjects. When we conducted experiments using a leave-one-subject-out (LOSO) approach, which ensures the same subject does not appear in both the training and test datasets, we achieved an accuracy of 79.4\%.

Furthermore, we compared our model with two recent studies by Santos et al.\cite{santhoshkumar2020vision} and Tahghighi et al.\cite{vua}. For this comparison, we adopted their methodology, allowing the same subjects to appear in both the training and test data. It's worth noting that this differs from the previously mentioned leave-one-subject-out (LOSO) approach. Our model achieved an accuracy of 99.02\%, which outperforms the closest reported result by Tahghighi et al.~\cite{vua} by a margin of 0.22\%. The results of our experiments, along with those from the two aforementioned studies, are presented in Table~\ref{multi2}.

The first approach yielded lower accuracy than the second one. This is largely because in the Leave-One-Subject-Out (LOSO) method, the model cannot memorize specific instances or characteristics of individual subjects. Instead, it must depend on learning patterns and features that can be generalized across different subjects. Moreover, it's required to perform effectively on subjects it hasn't encountered during training.





\begin{table}[]
    \centering
    \begin{tabular}{|c|c|c|}
    \hline
     \textbf{Author} & \textbf{Year}&\textbf{Accuracy (\%)}  \\
        \hline
       
        Santosh et al.~\cite{santhoshkumar2020vision}&2020&95.9\\
         Tahghighi et al.~\cite{vua}&2021 & 98.8\\ 
        \textbf{\name} &\textbf{2023}&\textbf{99.02}\\
        \hline
    \end{tabular}
    \caption{Comparison between our proposed \name multimodal emotion recognition and other recent works on the GEMEP dataset.}
    \label{multi2}
\end{table}

\subsection{Ablation Experiment}
We analyze the accuracy of each module of \name in a stand-alone manner, and in various combinations with other modules. Tests are performed with the datasets listed in table~\ref{datasets}. The resulting outputs from these experiments are shown in table~\ref{ablation}. We can see the face is most the effective source of information for emotion recognition. Adding background context improves accuracy but its effect reduces when more modes are used as multiple modes already give a good picture of the subject's emotion. Maximum accuracy is achieved when we use all three modes and contexts. Therefore, to achieve maximum accuracy we should always strive for using all the available data modes. 

\begin{table}[]
\centering
\begin{tabulary}{1.0\columnwidth}{|c|c|c|c|c|}
\hline
\textbf{F}&\textbf{P}&\textbf{G}&\textbf{ Accuracy without C (\%)} &\textbf{Accuracy with C (\%)}\\
\hline
\checkmark&&&68.1&69.7\\
\hline
&\checkmark&&42.3&45.6\\
\hline
&&\checkmark&34.8&37.2\\
\hline
\checkmark&\checkmark&&70.1&71.05\\
\hline
\checkmark&&\checkmark&69.71&70.9\\
\hline
\checkmark&\checkmark&\checkmark&71.5&71.82\\
\hline
\end{tabulary}
\caption{Ablation study: here, F, P, G and C stand for face, posture, gait and context respectively.}
\label{ablation}
\end{table}

\subsection{Computational Cost}
We measure the average processing time it takes to identify emotion using each module individually and in a multimodal manner. For better results, we run each of them separately to identify emotions on the laptop computer without any GPU acceleration. We see facial emotion recognition is the fastest while gait recognition is the slowest among the three modalities.

\begin{table}[]
\centering

\begin{tabulary}{1.0\columnwidth}{|c|c|c|c|}
\hline
\textbf{Face} &\textbf{Context}&\textbf{Posture}&\textbf{Gait}  \\ 
 \hline
 1&0.92&1.18&1.63\\

\hline
\end{tabulary}
\caption{Average time taken to classify emotion by each module in comparison to the face module only.}
\label{cost}
\end{table}

From table~\ref{cost} it is apparent that posture and gait take more processing time compared to the face module. However, in real-life situations such as a CCTV feed, a clear face may not be always available, and therefore, it is necessary to make use of these other modalities. Adding these modalities not only increases the accuracy but also the reliability of the system albeit increased computational cost. The trade-off can be decided based on specific system requirements.

\subsection{Situational Knowledge and Explanation Generation}\label{exp}

\begin{figure}[]
\centering
   \includegraphics[width=.6\linewidth]{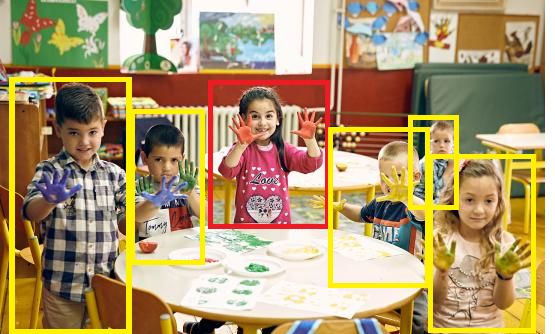}
   \caption{Explanation generation.}
   \label{kinder}
\vspace{-0.1in}
\end{figure}

Besides determining emotion we also provide a guideline to generate an explanation for that result. Our classification module generates emotion classification while our situational knowledge and explanation generation module provides human explainable reasoning by creating an idea of the situation around the subject. Individual modules tell us what information is available from the face, posture and gait cues. For instance, the subject in figure~\ref{kinder} red bounding box has a smiling face and open posture which indicates happiness. By extracting location type, location attributes and ANP information we create our situational knowledge which further enhances this reasoning. In the case of figure~\ref{kinder}, place category output is a classroom and the top three ANPs are early\_childhood, innoncent\_smile and creative\_kids. By combining all these a human understandable explanation can be constructed as "the subject is a child in a classroom doing creative work and smiling, has a happy facial expression and a body posture showing happiness". 

Using emotion obtained from other subjects in yellow boxes in the photo and calculating the average emotion score described in section~\ref{avgEmo} we find the happiness score is highest (0.80) in the leading position followed by neutral (0.2). So we can conclude this is a happy environment which bolsters our model's output of the kid in the red box to be happy.

This is the first effort for explainable emotion classification using multiple data modalities as per our knowledge. However, our explanation generation still requires further work. Therefore, we are planning to explore explanation generation in a more detailed manner in our future works.

\subsection{Anomaly Detection}\label{anomaly}
Anomaly detection is a possible use for multimodal recognition. An anomaly can be defined as a situation when a person intentionally tries to fake emotion to evade the system. Trying to fake an emotion most of the time results in a mismatch among the modalities as changing emotional expressions in face, posture and gesture require significant effort and skill.  The subject may be smiling while his body does not reflect happiness as such we are likely to get different results in the individual modules. If the modules differ beyond a certain threshold ($t_a$) in their output we tag that as an anomaly candidate. What to do with these anomaly data samples is a research question. Possible action can be further verification by a human in the loop and adaption of the model in a continuous learning fashion.  However, we keep this part as future work. 

\section{Conclusion and future works}\label{cfw}
In this paper, we present EMERSK, a modular system for emotion recognition from multimodal data. \name uses state of the art deep learning techniques to extract deep features from facial expression, posture, gait and background for emotion recognition. Moreover, it constructs situational knowledge from the place categories, ANPs and average emotion scores. We show how this situational knowledge can be useful to generate an explanation for the classification output. From the results of different experiments in several benchmark datasets, we report that \name has improved emotion recognition performance. In future, we plan to explore other modalities besides those used in this work. We also plan to automate the explanation generation process. Finally, we want to further explore anomaly detection and novelty in emotion recognition.


\bibliographystyle{IEEEtran}
\bibliography{main}

\vskip -2\baselineskip plus -1fil

\begin{IEEEbiography}[{\includegraphics[width=1in,height=1.25in,clip,keepaspectratio]{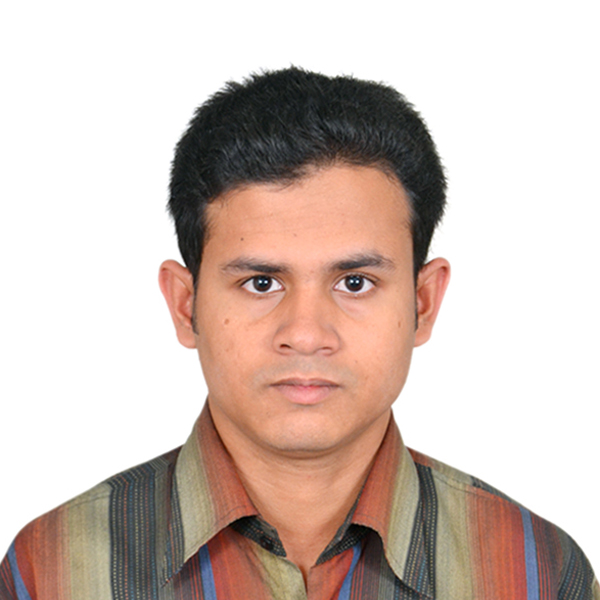}}]{Mijanur Palash}
received a BS degree in electrical and electronic engineering from Bangladesh University of Engineering and Technology and an MS in electrical and computer engineering from Southern Illinois University Carbondale, IL. Currently, he is working toward PhD in computer science at Purdue University, Indiana, USA. His research interests include multi-modal deep learning. 
\end{IEEEbiography}

\vskip -3\baselineskip plus -1fil

\begin{IEEEbiography}[{\includegraphics[width=1in,height=1.25in,clip,keepaspectratio]{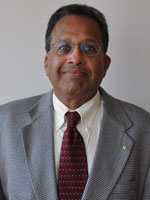}}]{Bharat Bhargava}
is a professor in the Department of Computer Science at Purdue University. He is a fellow of the Institute of Electrical and Electronics Engineers. In 1999, he received the IEEE technical achievement award. He is the founder of the IEEE Symposium on Reliable and Distributed Systems, the IEEE conference on Digital Library, and the ACM Conference on Information and Knowledge Management. He received his PhD in Electrical engineering from Purdue University in 1974. His current research interest includes intelligent autonomous system, data analytics and machine learning. 
\end{IEEEbiography}

\end{document}